\documentclass{article} 
\usepackage{iclr2027_conference,times}

\usepackage{amsmath,amsfonts,bm}

\def\eqref#1{equation~\ref{#1}}

\def\1{\bm{1}}

\DeclareMathAlphabet{\mathsfit}{\encodingdefault}{\sfdefault}{m}{sl}
\SetMathAlphabet{\mathsfit}{bold}{\encodingdefault}{\sfdefault}{bx}{n}

\usepackage{amsmath,amssymb}
\usepackage{arydshln}       
\usepackage{hyperref}
\usepackage{url}
\usepackage{booktabs}						
\usepackage{multirow}						
\usepackage{amsfonts}						
\usepackage{graphicx}						
\usepackage{duckuments}						
\usepackage{wrapfig}
\usepackage{ulem}
\usepackage{soul}
\usepackage{xcolor}
\usepackage{caption}

\usepackage{enumitem}
\setlist{topsep=2pt,itemsep=0pt,parsep=0pt}

\newcommand{\taut}{t}

\definecolor{BSUb}{RGB}{0,50,160}     
\definecolor{BSUb2}{RGB}{0,30,96}     
\definecolor{BSUb3}{RGB}{0,114,206}   
\definecolor{BSUo}{RGB}{250,70,22}    
\definecolor{BSUo2}{RGB}{255,130,0}   

\title{BrainDyn: A Sheaf Neural ODE for Forecasting and Generating Brain Dynamics}

\author{Siddharth Viswanath$^{1*}$, Panayiotis Ketonis$^{1*}$, Cassidy All$^{2*}$, William Dai$^{1*}$, \\
\textbf{Gabriel Escobedo}$^{1*}$, \textbf{Christine Gao}$^{3}$, \textbf{Jeffrey Lee}$^{4}$, \textbf{João Felipe Rocha}$^{1}$, \textbf{Andrew J Steindl}$^{1}$, \\
\textbf{Ian Adelstein}$^{1}$, \textbf{Michael Perlmutter}$^{5}$, \textbf{Dhananjay Bhaskar}$^{6}$, \textbf{Smita Krishnaswamy}$^{1}$ \\[4pt]
$^{1}$Yale University, USA \quad
$^{2}$University of Colorado Boulder, USA \\
$^{3}$University of Michigan Ann Arbor, USA \quad
$^{4}$Brown University, USA \\
$^{5}$Boise State University, USA \quad 
$^{6}$University of Wisconsin Madison, USA \\
\quad
*Equal Contribution \quad
Corresponding author: \texttt{smita.krishnaswamy@yale.edu}
}

\iclrfinalcopy 
\begin{document}

\maketitle

\begin{abstract}

\vspace{-4pt}

Brain activity unfolds as coordinated patterns across regions over time, and modeling these dynamics is a central goal in computational neuroscience. Because activity is distributed over connected brain regions, it is naturally represented on a graph whose nodes are brain regions or recording electrodes and whose edges capture relationships between these nodes, with message-passing modeling regional communication. However, standard message-passing networks place every node in a single shared feature space and typically aggregate messages from neighbors by scalar summation, which does not align with neurobiological evidence of brain communication. In order to model the heterogeneous nature of the brain, we introduce BrainDyn, a graph ordinary differential equation (ODE) network built on a cellular sheaf with dynamic restriction maps, recomputed from the forecasting  context window rather than fixed once, mediating region-to-region communication through learned linear transformations that drive the time evolution of neural activity. We evaluate BrainDyn on resting-state fMRI, EEG, and simulated activity from the NEST simulator. Across these modalities, we test multi-region forecasting and autoregressive generation of brain transients, and, using NEST, prediction of altered transients following controlled perturbations. BrainDyn outperforms competing models across modalities, while analysis of the learned features shows that they are expressive, interpretable, and biologically aligned.

\end{abstract}

\section{Introduction}

\vspace{-4pt}

A long-standing goal in computational neuroscience is a \textit{generative} model of brain activity, one that generates transient brain activity faithfully such that neuronal systems can be simulated in a data-driven way. Such models can create \textit{in silico} predictions before extensive experimentation: forecasting how activity will unfold and predicting responses to targeted perturbations~\cite{ramezanian2022generative, Markram2006}. 

Existing attempts at these tasks use modeling architectures that fall short in distinct ways. Standard sequence models, including recurrent networks and transformers, can capture temporal dependencies but are agnostic to functional brain organization; their predictions often cannot be attributed to specific brain regions. Graph neural networks (GNNs) offer a more natural fit, with parcellated regions, EEG channels, or individual neurons as nodes and anatomical or functional relationships as edges~\citep{luo2024graph, he2024spatiotemporal}. Yet standard message-passing architectures carry their own critical limitation: all nodes communicate in a single shared feature space, with aggregation typically performed by scalar summation or averaging~\citep{li2021braingnn, mohammadi2024graph}. In the brain, this is questionable since distinct regions differ systematically in the geometry and content of their representations. Furthermore, communication between regions is structured~\citep{lin2024topology, dubreuil2022role, semedo2019cortical, churchland2012neural}: regions encode information in distinct representational geometries~\citep{lin2024topology}, and communication between regions encoding different representations can be mediated by defined geometric transformations~\citep{semedo2019cortical}.
Moreover, standard message-passing oversmooths the heterogeneous, region-specific dynamics that characterize neural activity~\citep{rusch2023survey, wu2023demystifying}. We discuss additional related works in Appendix~\ref{app:related_works}.

We take inspiration from a cellular sheaf -- a mathematical structure that equips each edge with a linear map, called a restriction map. This structure has been utilized in sheaf neural networks~\citep{hansen2020sheaf, barbero2022sheaf, bodnar2022sheaf} that generalize standard GNNs by allowing each connection to implement a reversible transformation of the features into a common edge space, before diffusing into the space of neighboring nodes. Inspired by sheaf neural networks, we introduce \textbf{BrainDyn}, a neural sheaf ODE network with time-varying edge restriction and several unique features, including positional and time embeddings, implemented to allow learning of region-specific communication characteristics and time-varying dynamics. 

In \textbf{BrainDyn}, each node and edge is assigned a vector space, and an LSTM accumulates each region's recent history into its node space. Learnable and time-varying restriction maps are computed from the hidden features of the two nodes adjacent to each edge, allowing for restriction map modulation, a generalization of edge attention. Because these maps are output by MLPs parameterized by the current node states, they vary with time: different restriction maps are operational in different nodes states. At each timepoint, the restriction maps transform node features into edge-specific shared spaces and assemble a sheaf Laplacian. Message passing using this Laplacian models inter-region communication as structured, edge-specific transformations rather than scalar mixing. At each node, future transients are predicted by a time-parameterized derivative network which takes as input the aggregated sheaf message, as well as a positional encoding (indicating brain region) and a sinusoidal time embedding (indicating the temporal position in the transient). This time-dependent derivative is then passed to an ODE solver to integrate and forecast neural dynamics. 

We perform NEST simulations with a known ground-truth network of interacting neurons, allowing us to evaluate BrainDyn's ability to perform \textit{in silico} perturbations by showing that BrainDyn accurately models altered brain transients in response to controlled interventions. On the fMRI and EEG datasets, we evaluate BrainDyn's ability to forecast real-world resting-state activity at both short and long horizons, with the latter evaluated through autoregressive generation. Analysis of the learned model shows that the restriction maps selectively modulate connections and that sheaf-based message passing produces region-specific updates concentrated on specific sub-networks. Together, these results establish BrainDyn as a principled and expressive model of time-varying brain activity.

To summarize, our main contributions are:

\begin{enumerate}[topsep=2pt,itemsep=0pt,parsep=0pt]
\item \textbf{BrainDyn}, a dynamic sheaf-based graph ODE network that learns neural dynamics and generates brain-like transients. 
\item \textbf{Dynamic and learnable restriction maps} as expressive models of neural communication.
\item \textbf{Learned-frequency time embedding}: Integration of a learnable-frequency sinusoidal encoding into the
ODE, introducing explicit time dependence into the vector field.
\end{enumerate}

\section{Background}
\label{sec:bgnd}

\vspace{-4pt}

We model brain activity as signals supported on a graph $G=(V,E)$, where each node corresponds to a brain region and edges represent spatial proximity. To capture structured and heterogeneous interactions between the regions, we represent these signals using a sheaf over the graph at every point in time. In practice, the sheaves we use in our methods will vary over time.  

A cellular sheaf $\mathcal{F}$ on an undirected graph $G=(V,E)$ assigns  a vector space called a \textit{stalk} to each node and each edge and relates these stalks via linear maps known as \textit{restriction maps}. Formally, it consists of:
\begin{itemize}[topsep=0pt,itemsep=0pt]
   \item A vector space $\mathcal{F}(i) \cong \mathbb{R}^{d_i}$ for each node $i \in V$
    \item A vector space $\mathcal{F}(e_{ij})\cong \mathbb{R}^{d_{e_{ij}}}$ for each edge $e_{ij} \in E$
    \item Restriction maps $\rho_{i\rightarrow e_{ij}}: \mathcal{F}(i)\rightarrow \mathcal{F}(e_{ij}), \quad \rho_{j\rightarrow e_{ij}}: \mathcal{F}(j)\rightarrow \mathcal{F}(e_{ij})$
    for each edge $e_{ij} \in E$.
\end{itemize}

In principle, the dimensions of the stalks $d_i$ and $d_{e_{ij}}$ could vary from vertex and from edge to edge. However, in this work, we will assume that there is a shared dimension for all the stalks, i.e., $d_i=d_{e_{ij}}=d$ for all $i$ and $j$.



In the context of sheafs, signals (i.e., functions) defined on the vertices and the edges are represented by \textit{cochains}, where $0$-cochains represent signals defined on the vertices and $1$-cochains represent signals defined on the edges. Formally, we define the space of 0-cochains, $C^0(G,\mathcal{F})$, to be the direct sum of all the $\mathcal{F}(i), i \in V$, and similarly define  the space of $1$-cochains, $C^1(G,\mathcal{F})$, as the direct sum of all the $\mathcal{F}(e_{ij}), e_{ij} \in E$. We will denote an element of $C^0(G,\mathcal{F})$ by $H$ and let $h_i$ be the canonical projection of $H$ onto $\mathcal{F}(i)$. Intuitively, we think of $H$ as a function (signal) from $V$ to $\mathbb{R}^d$ and interpret $h_i$ as $H(i)$, the node features for vertex $i$.

We define the co-boundary operator $\delta: C^0(G,\mathcal{F})\rightarrow C^1(G,\mathcal{F})$ by 
\begin{equation}\label{eqn: delta classic}
    \delta(H)(e_{ij}) = \rho_{i\rightarrow e_{ij}}h_i - \rho_{j\rightarrow e_{ij}}h_j
\end{equation}
and define the \textit{sheaf Laplacian} as $\mathcal{L}_{\mathcal{F}}=\delta^\top\delta$, where $\delta^\top:C^1(G,\mathcal{F})\rightarrow C^0(G,\mathcal{F})$ is the adjoint of $\delta$. Somewhat analogous to the traditional graph incidence matrix, $\delta$ measures the discrepancy between $h_i$ and $h_j$ when they are mapped into a shared vector space.  The sheaf Laplacian then aggregates these discrepancies back to the nodes. 

To see this, we note that $\mathcal{L}_{\mathcal{F}}$ may be represented by a block-structured matrix where its diagonal blocks are given by $\mathcal{L}_F(i,i)=\sum_{e_{ij}\in E}\rho_{i\rightarrow e_{ij}}^\top\rho_{i\rightarrow e_{ij}}h_i$, and its off diagonal blocks are given by $\mathcal{L}_{\mathcal{F}}(i,j)=-\rho_{i\rightarrow e_{ij}}^\top\rho_{j\rightarrow e_{ij}}h_i$ for $e_{ij}\in E$,  $\mathcal{L}_{\mathcal{F}}(i,j)=0$ otherwise \citep{hansen2019learning}. This allows us to write 


\begin{equation}\label{eqn: sheaf Laplacian classic}
    (\mathcal{L}_{\mathcal{F}}H)_i = \sum_{\{j:e_{ij} \in E\}}\rho_{i\rightarrow e_{ij}}^\top(\rho_{i\rightarrow e_{ij}}h_i - \rho_{j\rightarrow e_{ij}}h_j)= \sum_{\{j:e_{ij} \in E\}}\rho_{i\rightarrow e_{ij}}^\top\delta(H)(e_{ij}).
\end{equation}


Therefore, the sheaf Laplacian generalizes the classical graph Laplacian by incorporating edge-dependent transformations before computing differences. Intuitively, it measures how inconsistent node features are across the graph after being aligned with the restriction maps. When neighboring nodes are consistent under these transformations, the Laplacian vanishes. 
Notably, the use of edge-specific restriction maps will allow us to produce distinct behavior in different regions of the brain.

In our experiments, we use the symmetric normalized sheaf Laplacian
\begin{equation}
\label{eqn: normalized sheaf laplacian}
\mathcal{L}_\mathcal{F}^{\text{norm}} = D_{\mathcal{F}}^{-1/2} \mathcal{L}_\mathcal{F} D_\mathcal{F}^{-1/2},\end{equation} where $D_\mathcal{F}$ is the block-diagonal matrix whose $i$-th diagonal block is
$D_\mathcal{F}(i,i)=\deg(i)I_d$, corresponding to normalization by the node degree.

The normalized sheaf Laplacian preserves the same underlying geometric information as the unnormalized sheaf Laplacian while rescaling connections according to degree. As a result, message-passing based will not depend on the scale of the restriction maps.

\begin{figure}
    \centering
    \includegraphics[width=1.00\linewidth]{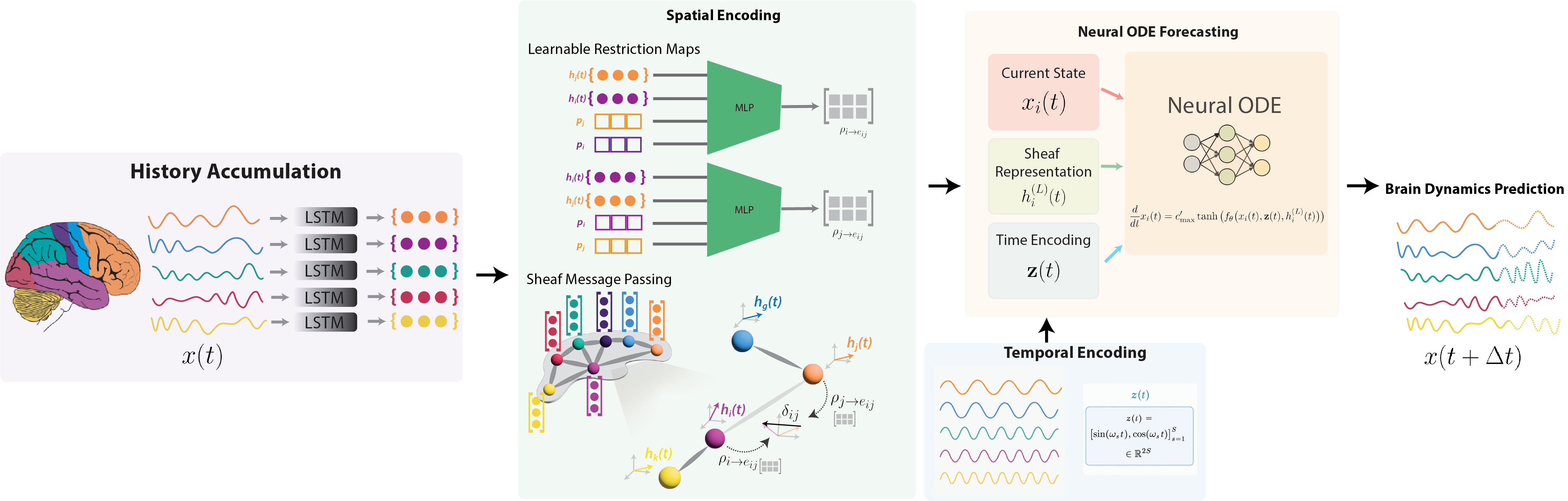}
    \vspace{-12pt}
    \caption{BrainDyn encodes each region's temporal history with an LSTM, and then uses learnable sheaf restriction maps to align and modulate interactions across brain regions. The resulting spatially encoded latent state is evolved with a neural ODE to predict future brain dynamics.}
    \label{fig:BrainDyn_Schematic}
    \vspace{-10pt}
\end{figure}

\section{BrainDyn: A Dynamic Sheaf-Based Graph ODE Network}
\label{headings}

\vspace{-4pt}

\textbf{BrainDyn} (as shown in Figure~\ref{fig:BrainDyn_Schematic}) is a dynamic sheaf-based graph ODE network designed for graph time-series forecasting and autoregressive generation. 
We assume that we are given a graph $G=(V,E)$, $|V|=N$, where each node $i \in V$ corresponds to a brain region. Edges are constructed via spatial proximity by computing pairwise Euclidean distances between the physical location of the nodes and connecting each node to its $k$ nearest neighbors. (In Appendix \ref{app:spatial_connectivity_results}, we perform an ablation comparing graph constructions based on spatial connectivity, functional connectivity, and Granger-causality and find that physical adjacency performs best.) We also assume that there is a continuous-time dynamical system $(x_i(t))_{t\geq 0}$ associated to each node and that we have access to a discrete time series of samples (i.e., observations) $x_i(t_0),x_i(t_1),x_i(t_2),\ldots,$ where 
$t_0=0$ and $t_{k+1}=t_k+\Delta t$ for a fixed step size $\Delta t$.


For time-series forecasting, we consider a context window consisting of the $T_c$ most recent observations  
\textcolor{black}{ $x_i({t_1^{\text{con}}}),\ldots,x_i(t^{\text{con}}_{T_c})$.} 
Our goal is to forecast the value of $x_i(t)$ for  $T_f$   future time-points, i.e., to predict  $x_i(t^{\text{fore}}_{1}),\ldots, x_i(t^{\text{fore}}_{\,T_f})$, using a method that utilizes both the temporal nature of the data and the graph structure $G$ (where $t_k^{\text{con}}$ denotes a time in the context window, $t_k^{\text{fore}}$ denotes a time in the forecast window). As detailed in Section \ref{sec:forecasting}, we will consider both short-term forecasting,
as well as long-term (autoregressive) generation.

Our framework consists of four components:
\begin{enumerate}[topsep=2pt,itemsep=0pt,parsep=0pt]

\item \textbf{History Accumulation:} 
To encode the recent temporal history of each node, we use an LSTM trained on observations in the context window. 

\item \textbf{Sheaf message passing via learnable and time-varying restriction maps:} 
To model the structured and heterogeneous communication between brain regions, we employ learnable, edge-specific, time-varying restriction maps $\rho_{i\rightarrow e_{ij}}(t)$ which align the node representations produced by the LSTM. This restriction map is time-varying because it is the output of an MLP that takes the current node features of nodes adjacent to the edge, i.e., it is akin to an attention mechanism which varies the transformation based on current node features. 

\item \textbf{Time and position embeddings} We model node identities using position embeddings as well as temporal information via a learned sinusoidal position embedding. 

\item \textbf{Forecasting via neural ODE}: 
To forecast future neural activity, we train a graph ODE, based on the output of the sheaf message passing and the learned time encoding. 
\end{enumerate}


\subsection{History Accumulation using LSTM}
\label{sec:temporal}
We first use an LSTM to encode the history of recent activity at each node. 
Specifically, for each node $i$, we consider a context vector 
\textcolor{black}{$[x_i(t^{\text{con}}_1), \ldots x_i(t^{\text{con}}_{\,T_C})]^\top$} consisting of observations at the $T_c$ most recent time points and define a hidden representation at time $t=t_{\,T_c}$ by 
%
%
\begin{equation*}
    h_i(t) = \texttt{LSTM}([x_i(t^{\text{con}}_1),\ldots x_i(t^{\text{con}}_{\,T_c})]^\top)\in\mathbb{R}^d,
\end{equation*}
where the $\texttt{LSTM}$ is a long short-term memory module that is shared across all $i$. We assume that $d$, the dimension of $h_i(t)$, is less than the length of the sequence $T_{c}$. Therefore, the LSTM  compresses complex temporal patterns into a latent dynamical state representing the recent neural activity at node  $i$ and time $t= t_{\,T_c}$. Below, in Section \ref{sec:restriction_map}, we will use  $h_i(t)$ as the stalk representation associated with node $i$ when applying methods based on the sheaf Laplacian. 

\subsection{Node feature-dependent Time-varying Learnable Restriction Maps}
\label{sec:restriction_map}

We train an MLP which  intakes the LSTM embeddings $h_{i}(t)$ and $h_{j}(t)$ for each edge $e_{ij}$ (defined as in Section \ref{sec:temporal})  and outputs the restriction maps at time $t$,  $\rho_{i\to e_{ij}}(t), \rho_{j\to e_{ij}}(t)\in \mathbb{R}^{d\times d}$. 

As additional input to the MLP, we also use learned per-node positional encodings \citep{devlin2019bert}; in our case, an 8-dimensional vector of learnable parameters per-node. These provide regional ``tags" that distinguish nodes and provide time-invariant regional identities (see Appendix~\ref{app:pe} for details). Explicitly, we fix $t=t^{\text{con}}_{\,T_c}$ to be the last point in the context window and compute:
%
\begin{equation}\label{eqn: MLP i first}
\rho_{i\to e_{ij}}(t) = \texttt{MLP}(h_i(t),h_j(t), p_i, p_j) = \texttt{MLP}([h_i^\top(t)|| h_j^\top(t)|| p^\top_i|| p^\top_j]^\top),
\end{equation}
where $p_i$ and $p_j$ are the positional encodings at nodes $i$ and $j$. We then define $\rho_{j\to e_{ij}}(t)$ by switching the order of concatenation, so that we satisfy the non-symmetry condition introduced in \citet{bodnar2022neural}:
\begin{equation}\label{eqn: MLP j first}
\rho_{j\to e_{ij}}(t)=\texttt{MLP}(h_j(t), h_i(t), p_i, p_j) = \texttt{MLP}([h_j^\top(t)|| h_i^\top(t)|| p^\top_j||p^\top_i]^
\top).
\end{equation}
In both cases, the MLP outputs a vector in $\mathbb{R}^{d^2}$, which is reshaped into the $d\times d$ restriction matrix.
We emphasize that we train a single MLP for use in both Eqn. \ref{eqn: MLP i first} and Eqn. \ref{eqn: MLP j first} and 
that the restriction maps are order dependent in the input arguments. We also note that the weights of this MLP are shared across edges and time points. Given the restriction maps, we obtain a learnable sheaf Laplacian $\mathcal{L}_F(t)$ and a learnable normalized sheaf Laplacian $\mathcal{L}_\mathcal{F}^\text{norm}(t)$ via Eqns.~\ref{eqn: sheaf Laplacian classic} and \ref{eqn: normalized sheaf laplacian}.

We note that the learned restriction maps $\rho_{i\to e_{ij}}(t), \rho_{j\to e_{ij}}(t)\in \mathbb{R}^{d\times d}$ are a generalization of neighbor-attention in graph attention networks~\citep{velickovic2018graph}. However, instead of the attention value being a scalar, the entire vector-space structure of each stalk is learned to modulate the flow of information at time $t$.

\subsection{Sheaf Message Passing}
\label{sec:sheaf_msg_passing}

We use the learnable sheaf Laplacian to perform iterative message passing and propagate information between brain regions based on the learned restriction maps. We again fix $t=t^{\text{con}}_{\,T_c}$ to be the last point in the context window, and we let $H^{(0)}(t) = H(t)$ denote the stacked node stalk representations given by 
\begin{equation*}
H(t) =
\begin{bmatrix}
h^\top_1(t), & h^\top_2(t), & \cdots &, h^\top_N(t)
\end{bmatrix}^\top
\in \mathbb{R}^{N\times d}.
\end{equation*}
We then perform $L$ rounds of sheaf message passing according to 
\begin{equation}\label{eqn: sheaf MP}
H^{(l)}(t) = (\mathcal{I} - \mathcal{L}^\text{norm}_{\mathcal{F}}(t))H^{(l-1)}(t), \quad l = 1, \dots, L,
\end{equation}
where $\mathcal{I}$ denotes the identity operator and $\mathcal{L}_\mathcal{F}^\text{norm}(t)$ is the learnable normalized sheaf Laplacian obtained by inserting the learned restriction maps defined in Section \ref{sec:restriction_map} into Eqn.~\ref{eqn: normalized sheaf laplacian} at time $t$. Each round of message passing  
 propagates information in 
 edge-specific vector spaces, i.e., the edge stalks. 
 The primary difference between this and standard graph-based message passing is that it allows each edge to apply a distinct learnable transformation prior to communication, enabling heterogeneous interactions between brain regions.
The resulting representation after $L$ rounds of message passing is denoted by $H^{(L)}(t)$, with $h_i^{(L)}(t)$ corresponding to the representation of node $i$.  Analogous to \citet{chamberlain2021grand}, we note that this message passing operation, Eqn. \ref{eqn: sheaf MP}, can be viewed as a forward-Euler discretization of the heat equation
$$
(\partial_t-\mathcal{L}_\mathcal{F}^\text{norm})H=0
$$
associated to the normalized sheaf Laplacian. Therefore, it can be interpreted as a graph diffusion. Alternatively, from the perspective of graph signal processing \citep{ortega2018graph,leus2023graph}, it can be viewed as a low-pass filtering.

\subsection{Sinusoidal Time Embedding for Non-Autonomous Dynamics }\label{sec: time embedding}


The sheaf message passing representation $h_i^{(L)}(t)$ encodes recent activity history of node $i$ together with information propagated from neighboring regions. This representation is computed once from the context window and held fixed during each forecast interval. However, brain dynamics evolve continuously with respect to time. Therefore, to introduce further time dependence, we compute a sinusoidal time encoding $\mathbf{z}(t)$. 

We let $d_t$ denote the time-embedding dimension, which we assume to be even, $d_t=2S$, and  define: 
\begin{equation*}
    \mathbf{z}(t)\hspace{-.03in} =\hspace{-.03in} \left[\sin\hspace{-.025in} \big(\omega_0(t-t_1^{\text{con}})\big),\dots, \sin\hspace{-.025in}\big(\omega_{S-1}(t-t_1^{\text{con}})\big), \cos\hspace{-.025in}\big(\omega_0(t-t_1^{\text{con}})\big),\dots, \cos\hspace{-.025in}\big(\omega_{S-1}(t-t_1^{\text{con}})\big)\right]^\top.
\end{equation*}
The  $\omega_k$ are frequencies   given by 
\begin{equation*}
    \omega_k \;=\; 2\pi\, c_{\max}\,\sigma(\eta_k),\quad k=0,\ldots,S-1,
\end{equation*}
where $\eta_k$ is a learnable parameter which is initialized so that $\omega_k = P^{-k/S}$ at the start of training. Here $P$ is a   maximum period, chosen based on the data,  $\sigma$ is the logistic function, and $c_\text{max}$ is an upper bound on  $\omega_k/2\pi$ in order to ensure that learned frequencies can always be resolved numerically. In practice, this constraint was never binding for any dataset.  We note that the encoding is global, and informs every node identically. The geometric-sequence initialization of $\omega_k$ encodes multiple timescales, allowing it to capture both fast and slow drift across each forecast window. 

\subsection{Modeling Neural Dynamics with a Neural ODE}

Neural activity is driven by both intrinsic dynamics within each brain region and interactions with other regions. These interactions are not static, but rather evolve over time and  depend on the current state of the system as well as ongoing coordination between regions.  

Therefore, we consider the following Neural ODE:
\begin{equation}\label{eqn: neural ode with the latent space}
    \frac{d}{d \taut} 
    x_i(t)= c'_\text{max}\tanh \big(f_\theta\big(x_i(t), \mathbf{z}(t), h^{(L)}_i(t)\big) \big), 
\end{equation}

where $f_\theta$ is parameterized as an MLP, $\mathbf{z}(t)$ is the time embedding introduced in Section \ref{sec: time embedding}, and $c'_\text{max} >0$ is a fixed constant which bounds the learned dynamics in order to ensure that they do not diverge. In practice, this bound was not active for any dataset.

\subsection{Forecasting Scenarios and Autoregressive Training}
\label{sec:forecasting}
Our goal is to forecast future neural activity,  given observations in the context window. Formally,  letting $x(t)=[x_1(t),\ldots, x_N(t)]^\top\in\mathbb{R}^N$ denote the global state of the system at some time $t$, we aim to use observations in the context window $x(t^{\text{con}}_1),\ldots,x(t^{\text{con}}_{\,T_c})$ to predict  $x(t^{\text{fore}}_1),\ldots,x(t^{\text{fore}}_{\,T_f})$, where $t^{\text{fore}}_1,\ldots,t^{\text{fore}}_{\,T_f}$ is a set of future time points (with $t^{\text{fore}}_1>t^{\text{con}}_{\,T_c}$). In our experiments, we will set $T_c=32$ and $T_f=8$ and will consider both short-term forecasting and long-term autoregressive generation (extensions to other values of $T_c$ and $T_f$ are straightforward).


\noindent\textbf{Short-term forecasting.} In the short-term scenario, we assume that the forecast window is immediately after the context window, i.e.,  $t^{\text{fore}}_1=t^{\text{con}}_{T_C} +\Delta t$. This allows us to directly apply BrainDyn by solving the ODE Eqn.~\ref{eqn: neural ode with the latent space}  using a fourth-order Runge-Kutta method, and then we evaluate the solution at the time points $t_k^{\text{fore}}$ in the forecast window.

  

\noindent\textbf{Long-term forecasting.} In the long-term scenario, we use an autoregressively trained model where we iteratively make short-term forecasts until we reach the designated forecast window. In our experiments, we set the forecast window to start $4T_f+1=33$ steps after the end of the context window, i.e., $t^{\text{fore}}_1=t^{\text{con}}_{T_C} +33\Delta t$. 
%
 We first apply BrainDyn as in the short-term forecasting scenario to predict the value of $x(t)$ at the next $T_f=8$ timepoints beyond the present $t = t^{\text{con}}_{T_c}$. 
We then shift our prediction window forward by eight time steps, inserting our predictions from the previous step as needed and use BrainDyn to predict the next 8 values. That is, if before we had $x(t_{\,T_c}^{\text{con}})=x(t_k)$ and our model input $[x(t_{k+1-T_f}),\ldots x(t_{k})]^\top$, it now inputs $[(x(t_{k+9-T_f}),\ldots x(t_{k}),{x}(t_{k+1}),\ldots{x}(t_{k+8})]^\top$ (where the last 8 inputs are based on our predictions from the first step). We repeat this process iteratively until we finally predict the values of $x(t)$ for $t$ within the prediction window.  Computational complexity of BrainDyn is discussed in Appendix~\ref{app:cc}. 


\section{Experiments}
\label{sec:experiments}

\vspace{-4pt}

We compare the performance of BrainDyn\footnote{The code is available at: \url{https://anonymous.4open.science/r/BrainDyn}}  
 against a variety of baseline models on three data sets:
 (i) resting-state functional magnetic resonance imaging (fMRI) data obtained from the Philadelphia Neurodevelopmental Cohort (PNC)~\citep{SATTERTHWAITE2014544} parcellated on the Schaefer 400 atlas; (ii) Resting-state EEG data  obtained from the MPI Leipzig Mind-Brain-Body database, specifically the Leipzig Study for Mind-Body-Emotion Interactions or LEMON~\citep{lemon}, a 62-channel scalp EEG study; and  (iii) synthetic neural activity generated using NEST~\citep{gewaltig2007nest}. We provide more information on these datasets  in Appendix~\ref{app:datasets}.


We utilize a 70/20/10 train/validation/test split. To prevent data leakage, we  ensure that all data associated to a given subject is contained in the same split. We repeat the training process five times (with different shuffles) and report the $\text{mean}\pm\text{std}$ over runs. 
We measure performance  in terms of mean squared error (MSE), mean absolute error (MAE), and Dynamic Time Warping (DTW). 
 We $z$-score normalize the node features  using statistics calculated per-node over all the train subjects' windows,   
 and train BrainDyn using the mean squared error loss function. We provide further  details on our experimental setup and on our performance metrics  in Appendices \ref{app:exp} and \ref{app:metrics}.


\subsection{Forecasting Brain Dynamics on fMRI and EEG data}

We demonstrate the performance of BrainDyn on two tasks: short-term forecasting and long-term autoregressive generation. As baselines, we consider an autoregressive CNN-LSTM~\citep{agga2022cnnlstm, alshembari2025autoregressive}, transformer models (BIOT~\citep{yang2023biot} and BrainLM~\citep{ortega2024brainlm}), a dynamic graphical model (EvolveGCN~\citep{egcn} and ODEBrain~\citep{jia2026ODEBRAIN}), and a graph neural ODE model (RiTINI~\citep{bhaskar2024inferring}). Detailed description of the baselines is provided in Appendix~\ref{app:baselines}. In addition, we compare against ablated versions of BrainDyn in order to demonstrate the utility of the sheaf-based message passing, the LSTM, and the time embedding. In Appendix~\ref{app:spatial_connectivity_results}, we perform an ablation study in which, rather than constructing the graph using spatial proximity, we construct the graph using functional connectivity and Granger causality. 

\begin{figure}[h]
    \centering
    \vspace{-8pt}
    \includegraphics[clip, width=0.6\textwidth]{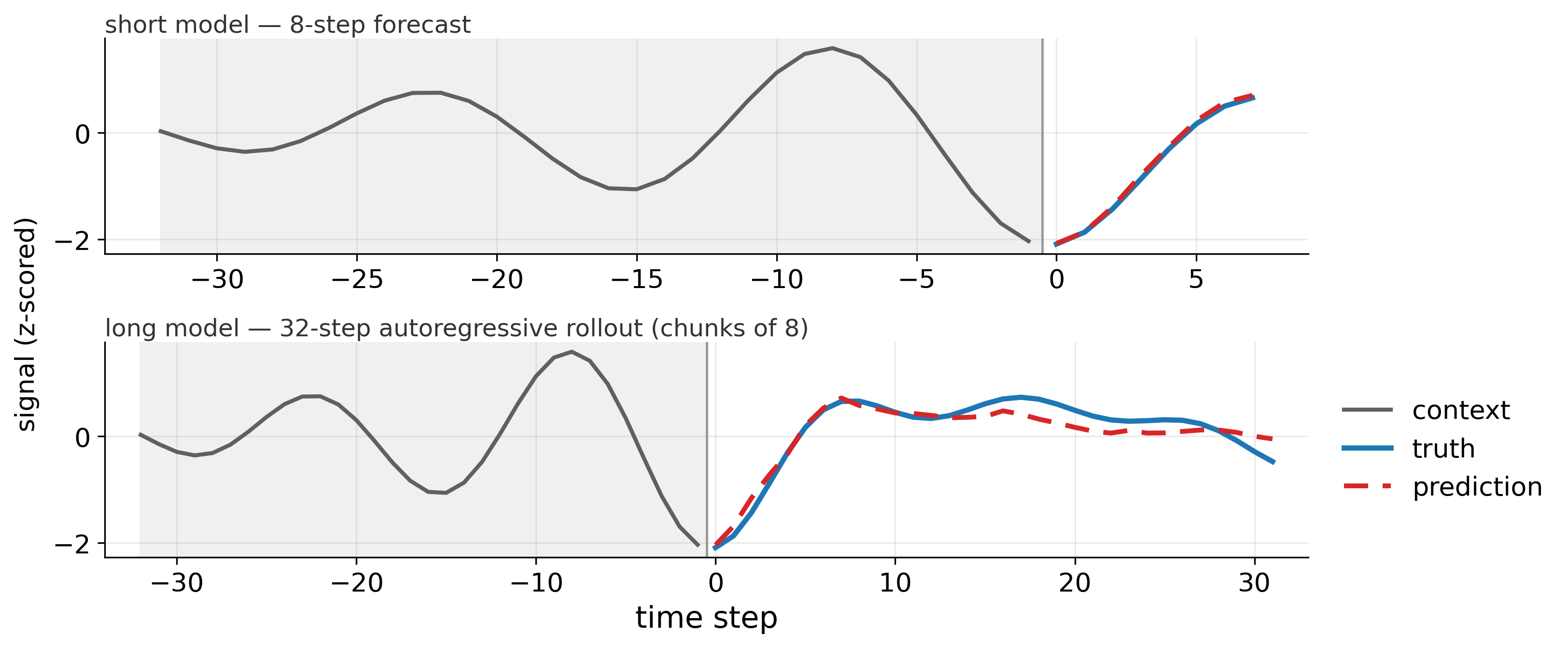}
    \vspace{-8pt}
    \caption{Ground truth vs predicted curve for fMRI short (top) and long autoregressive (bottom) prediction for a 32-context window in one fMRI region.}
    \label{fig:predictedcurves}
\end{figure}

\vspace{-8pt}

\noindent\textbf{fMRI.} Results for resting-state fMRI data are displayed in Table \ref{tab:forecasting_fmri}. In short-term forecasting, BrainDyn achieves the lowest error across all three metrics, reducing MSE by 45\% relative to the strongest baseline (CNN-LSTM) and DTW distance by 37\%. In long-term autoregressive generation, BrainDyn is again the top-performing method in all metrics, albeit by lower margins, reducing MSE by $5.5\%$ and DTW by $2.8\%$ compared to the top-performing baseline CNN-LSTM.  We also see in Tables \ref{tab:graphcomparison_fmri} and \ref{tab:graphcomparison_eeg} (Appendix \ref{app:spatial_connectivity_results}) that all of the ablations of BrainDyn cause performance to diminish across all metrics. We show an example of model output in Figure \ref{fig:predictedcurves}.

\begin{table}
\centering
\vspace{-20pt}
\caption{Prediction of short-term and long-term horizons on resting-state fMRI dynamics on the PNC dataset~\citep{satterthwaite2014neuroimaging}. Metrics reported as mean $\pm$ std across 5 randomized train/test/validation splits. Lower is better. Best results are \textbf{bolded}; second best are \underline{underlined}.}
\label{tab:forecasting_fmri}
\small
\setlength{\tabcolsep}{8pt}
\renewcommand{\arraystretch}{1.0}
\resizebox{\linewidth}{!}{%
\begin{tabular}{lcccccc}
\toprule
\multirow{2}{*}{\textbf{Method}} 
& \multicolumn{3}{c}{\textbf{Short forecasting}} & \multicolumn{3}{c}{\textbf{Long autoregressive generation}} \\
\cmidrule(lr){2-4} \cmidrule(lr){5-7}
& MSE~$\downarrow$ & MAE~$\downarrow$ & DTW~$\downarrow$ & MSE~$\downarrow$ & MAE~$\downarrow$ & DTW~$\downarrow$ \\
\midrule
CNN-LSTM & \underline{$0.0966 \pm 0.0212$} & \underline{$0.2110 \pm 0.0262$} & \underline{$0.0956 \pm 0.0108$} & $\underline{0.6746 \pm 0.0221}$ & $\underline{0.5725 \pm 0.0093}$ & $\underline{0.2713 \pm 0.0044}$ \\
BIOT & $1.0476 \pm 0.0496$ & $0.7425 \pm 0.0165$ & $0.3959 \pm 0.0088$ & $1.1481 \pm 0.0401$ & $0.7794 \pm 0.0133$ & $0.3951 \pm 0.0068$ \\
EvolveGCN & $0.9866 \pm 0.0535$ & $0.7209 \pm 0.0183$ & $0.3672 \pm 0.0092$ & $1.0082 \pm 0.0551$ & $0.7235 \pm 0.0174$ & $0.3821 \pm 0.0084$ \\
ODEBRAIN & $0.1385 \pm 0.0072$ & $0.2270 \pm 0.0077$ & $0.1029 \pm 0.0032$ & $0.7404 \pm 0.0252$ & $0.5891 \pm 0.0095$ & $0.3252 \pm 0.0037$ \\
BrainLM & $0.5869 \pm 0.4007$ & $0.4997 \pm 0.2172$ & $0.2475 \pm 0.1274$ & $1.1481 \pm 0.0401$ & $0.7793 \pm 0.0133$ & $0.3956 \pm 0.0067$ \\
RiTINI & $0.8940 \pm 0.0469$ & $0.6874 \pm 0.0171$ & $0.3569 \pm 0.0092$ & $1.0893 \pm 0.0432$ & $0.7589 \pm 0.0145$ & $0.3651 \pm 0.0078$ \\
\noalign{\vskip 3pt}
\cdashline{1-7}
\noalign{\vskip 3pt}
BrainDyn (Graph) & $0.7413 \pm 0.0522$ & $0.5905 \pm 0.0192$ & $0.2769 \pm 0.0091$ & $0.9136 \pm 0.0674$ & $0.6795 \pm 0.0206$ & $0.3467 \pm 0.0115$ \\
BrainDyn (w/o LSTM) & $0.7794\pm0.0502$ & $0.6216 \pm 0.0175$ & $0.2774 \pm 0.0078$ & $0.8682 \pm 0.0676$ & $0.6618 \pm 0.0212$ & $0.3313 \pm 0.0125$ \\
BrainDyn (w/o time emb.) & $0.2681 \pm 0.0362$ & $0.3477 \pm 0.0220$ & $0.1619 \pm 0.0109$ & $0.8177 \pm 0.0787$ & $0.6337 \pm 0.0276$ & $0.3337 \pm 0.0152$ \\
\noalign{\vskip 3pt}
\cdashline{1-7}
\noalign{\vskip 3pt}
\textbf{BrainDyn} & $\mathbf{0.0531\pm0.0091}$ & $\mathbf{0.1238\pm0.0121}$ & $\mathbf{0.0599\pm0.0054}$ & $\mathbf{0.6374 \pm 0.0465}$ & $\mathbf{0.5356 \pm 0.0157}$ & $\mathbf{0.2654 \pm 0.0084}$ \\
\bottomrule
\end{tabular}%
}
\end{table}

\noindent\textbf{EEG.} BrainDyn continues to perform well even in EEG as shown in Table \ref{tab:forecasting_eeg}. For short-term forecasting, it is the top-performing method across all three methods, beating the next best method ODE by $6.5\%$ on MSE.  In long-term autoregressive generation, BrainDyn is the top-performing method in MSE and MAE, while it is second to ODEBRAIN with a $19\%$ higher DTW. Overall, we note that BrainDyn has strong performance across both data modalities and metrics while other methods exhibit less consistent performance. Note that BrainLM uses 3-D coordinates as spatial embeddings in the fMRI setting; here, we use the EEG analog, 3-D sensor position on the scalp. We also see that ablations of BrainDyn generally cause performance to diminish, especially on the fMRI dataset.

\begin{table}[htbp]
\centering
\vspace{-10pt}
\caption{Prediction of short-term and long-term horizons on resting-state EEG dynamics on the LEMON dataset~\citep{lemon}. Metrics reported as mean $\pm$ std across 5 randomized train/test/validation splits. Lower is better. Best results are \textbf{bolded}; second best are \underline{underlined}.}
\label{tab:forecasting_eeg}
\small
\setlength{\tabcolsep}{8pt}
\renewcommand{\arraystretch}{1.0}
\resizebox{\linewidth}{!}{
\begin{tabular}{lcccccc}
\toprule
\multirow{2}{*}{\textbf{Method}} 
& \multicolumn{3}{c}{\textbf{Short forecasting}} & \multicolumn{3}{c}{\textbf{Long autoregressive generation}} \\
\cmidrule(lr){2-4} \cmidrule(lr){5-7}
& MSE~$\downarrow$ & MAE~$\downarrow$ & DTW~$\downarrow$ & MSE~$\downarrow$ & MAE~$\downarrow$ & DTW~$\downarrow$ \\
\midrule
CNN-LSTM & $0.6624 \pm 0.0995$ & $0.6016 \pm 0.0448$ & $0.2888 \pm 0.0209$ & $0.8713 \pm 0.1128$ & $0.6822 \pm 0.0364$ & $0.3488 \pm 0.0178$ \\
BIOT & $0.9652 \pm 0.1684$ & $0.7126 \pm 0.0601$ & $0.3799 \pm 0.0321$ & $0.9951 \pm 0.1639$ & $0.7225 \pm 0.0558$ & $0.3660 \pm 0.0280$ \\
EvolveGCN & $0.6870 \pm 0.1308$ & $0.5866 \pm 0.0541$ & $0.2874 \pm 0.0266$ & $0.9247 \pm 0.1682$ & $0.6912 \pm 0.0593$ & $0.3230 \pm 0.0279$ \\
ODEBRAIN & $\underline{0.4278 \pm 0.0788}$ & $\underline{0.4450 \pm 0.0337}$ & $\underline{0.2039 \pm 0.0153}$ & $\underline{0.7521 \pm 0.1329}$ & $\underline{0.6228 \pm 0.0519}$ & $\mathbf{0.2522 \pm 0.0197}$ \\
BrainLM & $0.9648 \pm 0.1881$ & $0.7125 \pm 0.0672$ & $0.3800 \pm 0.0359$ & $0.8993 \pm 0.2504$ & $0.6840 \pm 0.0911$ & $0.3185 \pm 0.0769$ \\
RiTINI & $0.9508 \pm 0.1858$ & $0.7077 \pm 0.0667$ & $0.3708 \pm 0.0350$ & $0.9837 \pm 0.1823$ & $0.7185 \pm 0.0624$ & $0.3469 \pm 0.0314$ \\
\noalign{\vskip 3pt}
\cdashline{1-7}
\noalign{\vskip 3pt}
BrainDyn (Graph) & $0.4867 \pm 0.0231$ & $0.4678 \pm 0.0140$ & $0.2141 \pm 0.0072$ & $0.7968 \pm 0.0732$ & $0.6314 \pm 0.0301$ & $0.3271 \pm 0.0151$ \\
BrainDyn (w/o LSTM) & $0.7214 \pm 0.0290$ & $0.5845 \pm 0.0166$ & $0.2742 \pm 0.0078$ & $0.9253 \pm 0.0926$ & $0.6819 \pm 0.0368$ & $0.3484 \pm 0.0194$ \\
BrainDyn (w/o time emb.) & $0.4981 \pm 0.0455$ & $0.4897 \pm 0.0258$ & $0.2366 \pm 0.0111$ & $0.7780 \pm 0.0688$ & $0.6099 \pm 0.0628$ & $0.3199 \pm 0.0137$ \\
\noalign{\vskip 3pt}
\cdashline{1-7}
\noalign{\vskip 3pt}
\textbf{BrainDyn} & $\mathbf{0.4001 \pm 0.0253}$ & $\mathbf{0.4209 \pm 0.0126}$ & $\mathbf{0.1905 \pm 0.0292}$ & $\mathbf{0.7199 \pm 0.0292}$ & $\mathbf{0.6005 \pm 0.0161}$ & $\underline{0.3008 \pm 0.0080}$ \\
\bottomrule
\end{tabular}
}
\end{table}

\subsection{Forecasting Perturbed Dynamics}
\label{sec:results_ood}

We test the ability of BrainDyn to generalize to interventions that were not observed during training. We used the NEST simulator~\citep{gewaltig2007nest} to generate activity from networks of 100 integrate-and-fire neurons with alpha-shaped postsynaptic currents. We generated 1,000 independent instantiations of this network, each with randomly sampled neuron positions and connected according to a distinct sparse directed small-world graph with 400 edges. The directed small-world graph is constructed through the following process: there is a shared spine of $400$ edges for every NEST graph, and then in each NEST graph, $10\%$ of the spine is rewired randomly. For consistency with the structural prior used in the fMRI and EEG datasets, we build the prior graph here with the spine of $400$ edges. All neurons in each network received the same shared Poisson background spike train at rate 1,000 Hz, and the recurrent synaptic weights were uniform (delay 1.5 ms). We binned spike trains into 10-millisecond windows and smoothed the result with a Gaussian kernel ($\sigma = 20$ ms) to obtain per-neuron firing-rate time series, which served as the observed signal for all models. To simulate perturbation, we silenced a single neuron over a contiguous window of random duration $D \in [240, 400]$~ms, beginning at a random onset time $t_0$ within the inner 80\% of each 2,000-ms simulation. This intervention directly alters the activity of the targeted neuron and propagates through the network according to the simulated connectivity. Full simulation parameters are provided in Appendix~\ref{app:datasets}.

We constructed context windows that straddle the perturbation onset, with 10\% of the context before onset and 10\% after onset. Thus, the input contains the perturbation onset and early network response, while the prediction window covers the subsequent perturbed trajectory. Critically, all models were trained only on unperturbed activity from held-out networks and evaluated on perturbed windows without further training. This setting tests whether the learned dynamics extrapolate to out-of-distribution activity patterns, a key property for an \textit{in silico} simulator of neural interventions. Like the fMRI and EEG experiments, the models trained on the NEST dataset had two training modes: training on either single-shot forecasting for short-term horizons or autoregressive forecasting for long-term horizons. However, unlike fMRI and EEG, we do not pass positional encodings into the restriction map MLP as the NEST nodes do not have a consistent identity between graphs in the same way as actual brain regions.

\begin{table}[!htbp]
\centering
\vspace{-8pt}
\caption{Prediction of short-term and long-term horizons on the synthetic NEST dataset of perturbed, out-of-distribution brain dynamics, with the structural graph prior and only 10\% of the 32-bin context window falling at or after the perturbation onset (3 of 32 bins); all models are trained and evaluated at this onset position. Metrics reported as mean $\pm$ std across 5 randomized train/test/validation splits. Lower is better. Best results are \textbf{bolded}; second best are \underline{underlined}.}
\label{tab:perturbed_dynamics_structural_frac01}
\small
\setlength{\tabcolsep}{8pt}
\renewcommand{\arraystretch}{1.1}
\resizebox{\linewidth}{!}{
\begin{tabular}{lcccccc}
\toprule
\multirow{2}{*}{\textbf{Method}}
& \multicolumn{3}{c}{\textbf{Short forecasting}} & \multicolumn{3}{c}{\textbf{Long autoregressive generation}} \\
\cmidrule(lr){2-4} \cmidrule(lr){5-7}
& MSE~$\downarrow$ & MAE~$\downarrow$ & DTW~$\downarrow$ & MSE~$\downarrow$ & MAE~$\downarrow$ & DTW~$\downarrow$ \\
\midrule
CNN-LSTM & $1.1932 \pm 0.0068$ & $0.8257 \pm 0.0031$ & $0.4404 \pm 0.0017$ & $1.1754 \pm 0.0048$ & $0.8299 \pm 0.0034$ & $0.4417 \pm 0.0022$ \\
BIOT & $1.1937 \pm 0.0069$ & $0.8261 \pm 0.0032$ & $0.4399 \pm 0.0016$ & $1.1888 \pm 0.0037$ & $0.8348 \pm 0.0020$ & $0.4343 \pm 0.0016$ \\
EvolveGCN & $1.1909 \pm 0.0064$ & $0.8253 \pm 0.0029$ & $0.4373 \pm 0.0015$ & $1.1738 \pm 0.0040$ & $0.8290 \pm 0.0030$ & $0.4413 \pm 0.0016$ \\
ODEBRAIN & $\underline{0.7197 \pm 0.0131}$ & $\underline{0.6063 \pm 0.0071}$ & $\underline{0.2858 \pm 0.0035}$ & $\underline{1.0505 \pm 0.0067}$ & $\underline{0.7701 \pm 0.0025}$ & $\underline{0.3988 \pm 0.0012}$ \\
RiTINI & $1.1819 \pm 0.0065$ & $0.8230 \pm 0.0029$ & $0.4363 \pm 0.0017$ & $1.1725 \pm 0.0034$ & $0.8276 \pm 0.0017$ & $0.4377 \pm 0.0007$ \\
\noalign{\vskip 3pt}
\cdashline{1-7}
\noalign{\vskip 3pt}
BrainDyn (Graph) & $0.8919 \pm 0.0075$ & $0.7051 \pm 0.0025$ & $0.3439 \pm 0.0012$ & $1.0989 \pm 0.0032$ & $0.7959 \pm 0.0013$ & $0.4145 \pm 0.0008$ \\
BrainDyn (w/o LSTM) & $0.8846 \pm 0.0082$ & $0.7041 \pm 0.0026$ & $0.3420 \pm 0.0011$ & $1.0994 \pm 0.0032$ & $0.7966 \pm 0.0011$ & $0.4155 \pm 0.0010$ \\
BrainDyn (w/o time emb.) & $0.7281 \pm 0.0309$ & $0.6330 \pm 0.0119$ & $0.3198 \pm 0.0045$ & $1.0488 \pm 0.0051$ & $0.7722 \pm 0.0020$ & $0.4072 \pm 0.0010$ \\
\noalign{\vskip 3pt}
\cdashline{1-7}
\noalign{\vskip 3pt}
\textbf{BrainDyn} & $\mathbf{0.6649 \pm 0.0109}$ & $\mathbf{0.5682 \pm 0.0053}$ & $\mathbf{0.2686 \pm 0.0029}$ & $\mathbf{1.0202 \pm 0.0043}$ & $\mathbf{0.7473 \pm 0.0025}$ & $\mathbf{0.3896 \pm 0.0014}$ \\
\bottomrule
\end{tabular}
}
\end{table}

Table~\ref{tab:perturbed_dynamics_structural_frac01} reports short-term and long-term forecasting performance under this perturbation protocol respectively. In both scenarios, BrainDyn is the top-performing method across all three performance metrics. Compared with ODEBRAIN, the strongest baseline in both the short-term and long-term, BrainDyn reduces MSE by 8\%~($0.7197$ to $0.6649$), MAE by 6\%~($0.6063$ to $0.5682$), and DTW distance by 6\%~($0.2858$ to $0.2686$) in the short term, and it reduces MSE by 3\%~($1.0505$ to $1.0202$), MAE by 3\%~($0.7701$ to $0.7473$), and DTW distance by 2\%~($0.3988$ to $0.3896$) in the long term.


We again see that ablating the sheaf structure, the time embedding, or the LSTM results in a drop in performance, further indicating that  each of these components improves BrainDyn's ability to extrapolate to perturbed neural dynamics.


\subsection{Understanding the Learned Restriction Maps}
\begin{figure}[h]
    \centering
    \vspace{-12pt}
    \includegraphics[clip, width=1\textwidth]{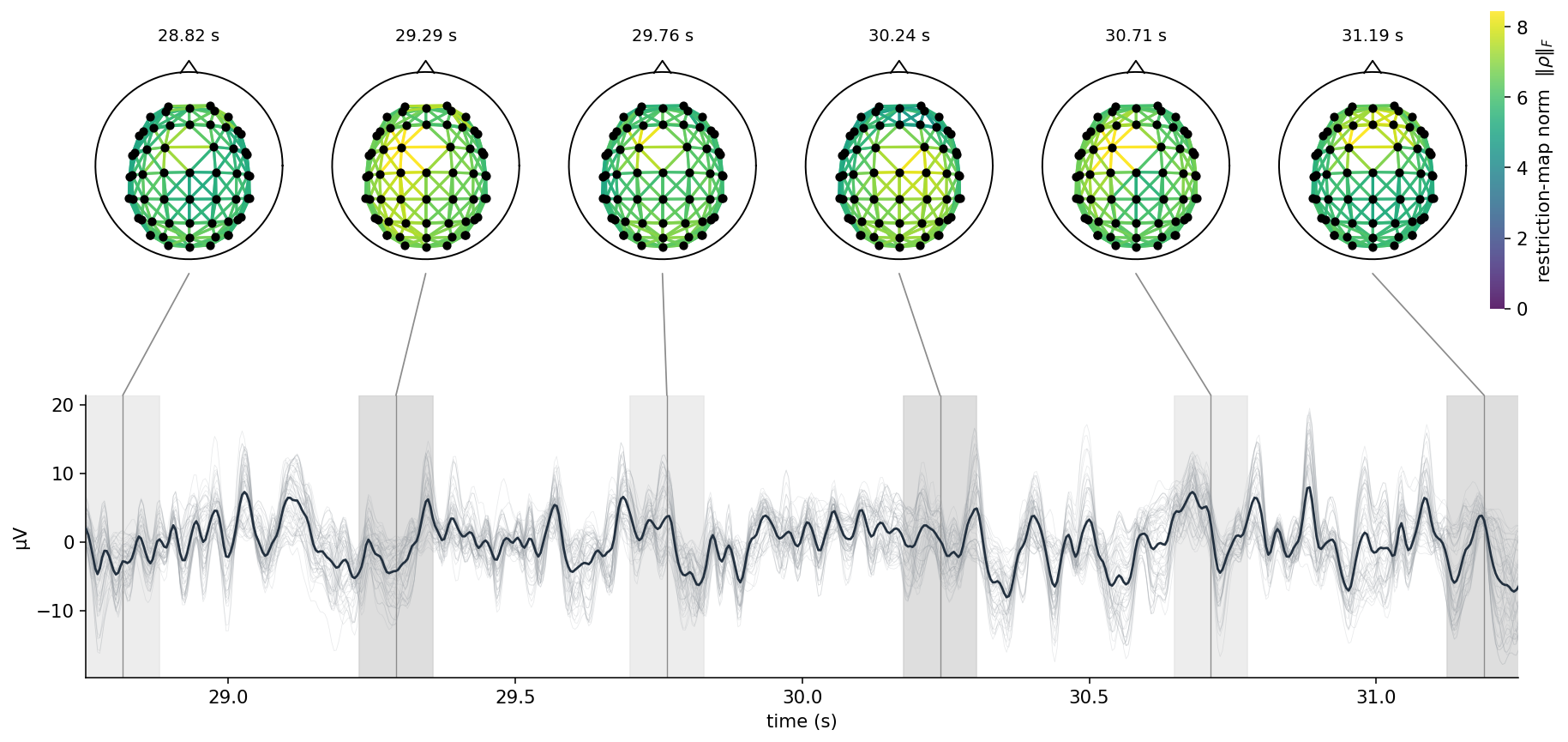}
    \vspace{-10pt}
    \caption{Highlighted windows of scalp-projected EEG with edges colored by the average Frobenius norm $\|\cdot\|_F$ of restriction maps in each time window. Edge norms vary across edges and time, consistent with brain region communication being dynamically modulated.  Thin gray lines: individual sensors over time; bold line: average signal across all sensors.}
    \label{fig:mapnorms}
\end{figure}

 BrainDyn learns structured interactions between brain regions through the restriction maps. 
 To highlight this, Figure~\ref{fig:mapnorms} illustrates the temporal and edge-wise variation of (the magnitude of) the learned restriction maps as quantified by the Frobenius norm. We see that the norm of the restriction map varies both over time and between brain regions. 
This  is consistent with neuroscientific findings that brain regions dynamically modulate  interactions
~\citep{sakalar2022neurogliaform, phan2024dynamic}. 

\begin{wrapfigure}{r}{0.42\textwidth}
    \centering
    \vspace{-16pt}
    \includegraphics[width=0.42\textwidth]{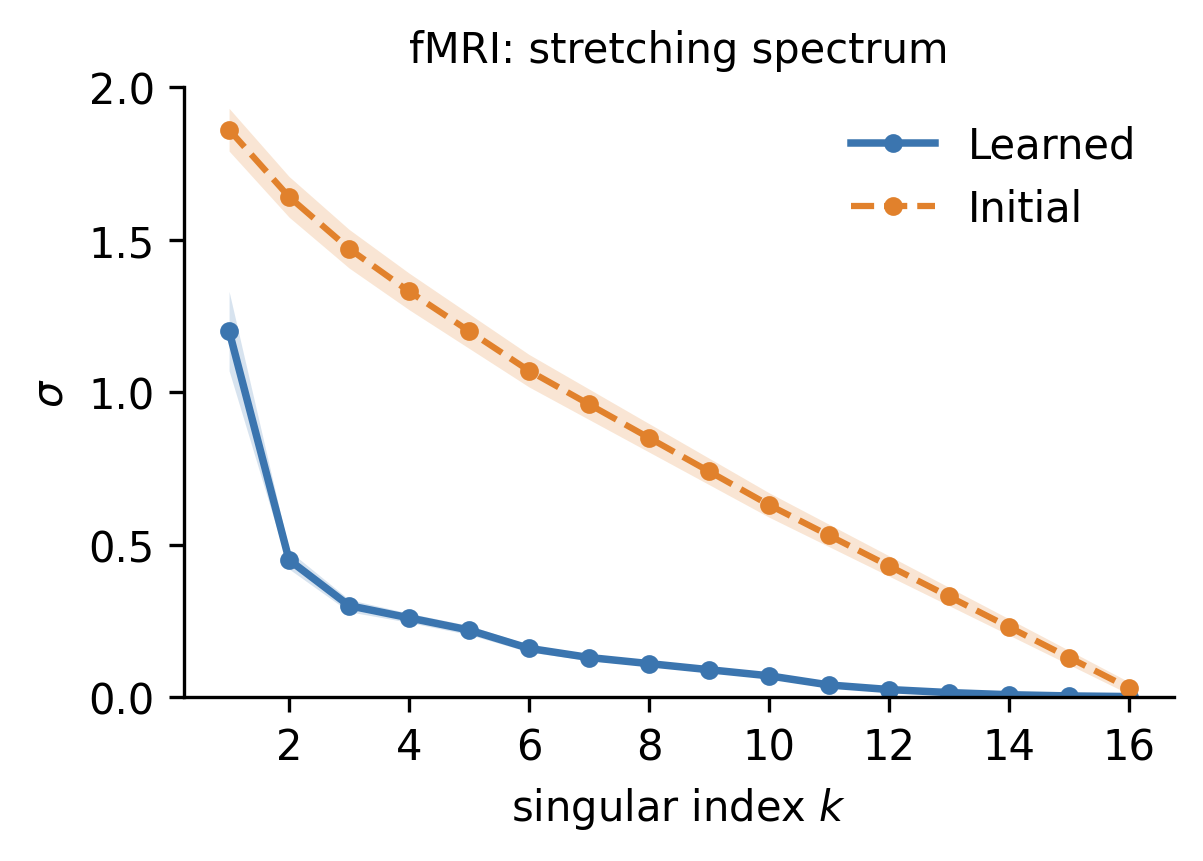}
    \vspace{-26pt}
    \caption{Singular values of the learned restriction maps compared to initial maps.}
    \label{fig:mapfactorization}
    \vspace{-10pt}
\end{wrapfigure}
The singular value decomposition factors the restriction maps into two orthogonal matrices $U$ and $V$, as well as a diagonal $\Sigma$, $\rho_{i\to e_{ij}}(t)=U \Sigma V^T$, providing a geometric interpretation of regional computation and communication for biological insights. We perform SVD on all of the maps and analyze the population. We find that the restriction maps are substantially changed from their initialization, with the mean singular value spectrum rescaled and reshaped (see Figure~\ref{fig:mapfactorization}). 
In particular, the singular values of learned maps exhibit rapid decay indicating an approximate low-rank structure, consistent with evidence that inter-areal neural communication is mediated through a low-dimensional communication subspace~\citep{semedo2019cortical}. Additional results for interpretability of the restriction maps are provided in Appendix~\ref{app:stretch-rotation}.

\section{Conclusion}
\label{sec:conclusion}

\vspace{-4pt}

We introduced BrainDyn, a dynamic (time-varying) sheaf-based graph neural ODE for modeling continuous-time neural dynamics. BrainDyn features several innovations over sheaf neural networks including 1) learnable time-varying restriction maps that generalize graph attention, 2) integrating sheaf-based networks with graph ODEs (which has not been done previously), and 3) temporal embeddings to learn non-autonomous dynamics.  We show that BrainDyn is able to effectively forecast fMRI and EEG data, as well as autoregressively generate long transients. Further, we use NEST simulations to demonstrate the capacity of BrainDyn to model unobserved perturbations. Overall this framework provides a mechanistically interpretable and expressive network for generating complex brain-like transients.  
Interesting avenues of future work include exploring the application of BrainDyn to larger and more diverse neural datasets, particularly in downstream, application-driven tasks, as well as developing methods for correlated time series in other domains.

\section*{Acknowledgments}

M.P. received funding from the National Science Foundation under grant number OIA-2242769. S.K. and
M.P. also acknowledge funding from NSF-DMS Grant No. 2327211. S.K acknowledges funding
from NSF CAREER award IIS-2047856. S.K also acknowledges funding from the Wu Tsai Institute
at Yale University.

\subsection*{AI usage statement}

We used Large Language Models (LLMs) for writing and coding. We used LLMs to implement parts of BrainDyn code. We have not used generative AI tools to generate synthetic datasets, help develop theoretical models or conceptual frameworks, formulate mathematical claims, provide ingredients for proving mathematical claims, propose or refine hypotheses, design research methodology, or clean datasets. Additionally, we used LLMs to draft, edit, and structure parts of the paper, to identify and summarize relevant literature, to format tables, and to create and edit software code. We have reviewed and revised all AI-assisted text, and reviewed and tested AI-assisted code and computed values.

\subsection*{Reproducibility statement}

Source code for BrainDyn, including the model, training and evaluation scripts, data preprocessing, and the NEST simulation pipeline, is available at an anonymous repository (\url{https://anonymous.4open.science/r/BrainDyn}), linked in Section~\ref{sec:experiments}. The model is fully specified in Section~\ref{headings}: the sheaf Laplacian and its degree normalization are defined in Section~\ref{sec:bgnd}, the LSTM history encoder in Section~\ref{sec:temporal}, the restriction map MLP and positional encodings in Section~\ref{sec:restriction_map}, sheaf message passing in Section~\ref{sec:sheaf_msg_passing}, the learned-frequency time embedding in Section~\ref{sec: time embedding}, and the short-term and autoregressive forecasting procedures in Section~\ref{sec:forecasting}. Appendix~\ref{app:datasets} describes the data sources and all preprocessing steps, covering the PNC fMRI derivatives (atlas, nuisance regression, windowing), LEMON EEG channel interpolation and windowing, per-node $z$-scoring, and the NEST network, simulation, and perturbation parameters. The subject-level 70/20/10 split and five-run protocol are described in Section~\ref{sec:experiments}, and the optimizer, learning-rate schedule, initialization, ODE solver, training length, batch size, hardware, and architecture hyperparameters are reported in Appendix~\ref{app:exp}. Evaluation metrics, including the DTW normalization, are defined in Appendix~\ref{app:metrics}, and the implementation and adaptation of each baseline to our setting are described in Appendix~\ref{app:baselines}. Appendix~\ref{app:spatial_connectivity_results} specifies the alternative graph constructions used in the ablation, Appendix~\ref{app:cc} gives the computational complexity.

\newpage

\bibliography{iclr2027_conference}
\bibliographystyle{iclr2027_conference}

\newpage

\appendix

\section{Extended Related Works}
\label{app:related_works}

\paragraph{Graph neural networks for brain signals.}
Graph neural networks provide a natural inductive bias for brain data, where electrodes, parcellated regions, or individual neurons can be modeled as nodes and anatomical or functional relationships as edges. Spectral and spatial GCN variants have been applied to fMRI and EEG alike, with message passing proving especially effective for seizure detection and classification~\citep{tang2022seizure}. Dynamic graph settings, where connectivity changes over time, have been addressed by architectures such as EvolveGCN~\citep{egcn}, which allows graph filter weights to evolve as a function of temporal context, and temporal graph network variants that jointly model recurrent state and graph structure~\citep{jia2026ODEBRAIN}. Despite these advances, standard message-passing architectures share a common assumption: all nodes communicate in the same feature space, with aggregation performed by scalar summation or averaging. In the brain, this assumption is questionable: distinct regions differ systematically in the dimensionality, geometry, and content of their representations. Repeated averaging across heterogeneous regions leads to over-smoothing, progressively homogenizing node states and erasing the regional specificity that is a defining characteristic of neural computation. These limitations motivate a richer interaction model.


\paragraph{Cellular sheaves and sheaf neural networks.}
A cellular sheaf over a graph assigns a vector space, termed a stalk, to each node and edge, together with linear restriction maps that relate neighboring stalks. The sheaf Laplacian, derived from these maps, generalizes the classical graph Laplacian by measuring inconsistency between node features only after they have been transformed into a shared edge space~\citep{hansen2020sheaf}. \citet{hansen2020sheaf} first proposes building neural networks around this structure, showing that sheaf convolutions generalize GCNs and naturally handle asymmetric and heterogeneous relational structure. Bodnar et al.~\citep{bodnar2022sheaf} provided a comprehensive theoretical treatment of Neural Sheaf Diffusion, establishing that non-trivial sheaves can prevent over-smoothing and accommodate heterophilic graphs where standard GNNs degrade. \citet{barbero2022sheaf} further connects sheaf learning to Riemannian geometry, proposing connection Laplacians derived from local tangent space alignment as a geometrically principled approach to constructing restriction maps. In aggregate, this body of work establishes that equipping each edge with a learnable linear transformation yields strictly more expressive message-passing processes than scalar aggregation. BrainDyn applies this insight to neural dynamics: restriction maps are learned per connection, allowing distinct brain regions to maintain separate representational geometries while interacting through structured, edge-specific transformations.



\paragraph{Neural ODEs and continuous-time dynamics on graphs.}
Neural ordinary differential equations parameterize the derivative of a hidden state with a neural network and integrate it using an adaptive numerical solver, enabling continuous-time modeling without fixed-step discretization~\citep{chen2018neural}. Rubanova et al.~\citep{rubanova2019latent} extended this to latent variable models for irregularly-sampled time series, where observations arrive at non-uniform intervals that recurrent architectures handle poorly. Graph neural ODEs further combine continuous-time latent dynamics with graph-structured message passing, allowing the vector field to depend on relational context~\citep{poli2019graph}. ODEBRAIN constructed a dual encoder and integrated them through a gated neural ODE field to forecast evolving EEG graph representations~\citep{jia2026ODEBRAIN}. RiTINI coupled space-time graph attention with a graph neural ODE to infer time-varying regulatory interaction graphs, demonstrating the value of continuous-time propagation for network-level dynamical inference~\citep{bhaskar2024inferring}. A shared limitation of these approaches is that all node representations live in a single, homogeneous latent space. BrainDyn departs from this by embedding the ODE vector field within a sheaf, so the interaction structure driving continuous-time evolution is itself geometrically heterogeneous and connection-specific.


\paragraph{Biosignal transformers and sequence models.}
Transformer-based architectures have emerged as strong general-purpose encoders for biosignals. BIOT~\citep{yang2023biot} tokenizes multichannel waveforms via short-time Fourier transforms and applies a Linformer over the resulting token sequence, enabling cross-dataset learning across heterogeneous EEG recording configurations. Related work on pretraining models has shown that self-supervised objectives over raw or frequency-domain representations yield representations that transfer to downstream clinical tasks. At a simpler level, hybrid convolutional-recurrent architectures capture local temporal features with CNNs before aggregating them with an LSTM, offering a compact baseline for multivariate forecasting~\citep{agga2022cnnlstm, alshembari2025autoregressive}. A common limitation of these sequence models is that they are largely agnostic to brain structure: channels or regions are treated as independent or fully-connected, without explicit modeling of anatomical or functional connectivity. This weakens their utility as \textit{in silico} simulators, where the organization of the brain graph should constrain how predictions propagate across the network.



\paragraph{Generative and foundation models for neural data.}

The ambition of building a generative model of the brain has a long history in computational neuroscience. The Virtual Brain project helped establish the idea that empirical connectivity can be combined with dynamical systems to simulate whole-brain activity, motivating \textit{in silico} brain models that can generate and forecast neural dynamics \citep{sanz2013virtual}. More recently, large-scale foundation models have brought a complementary data-driven approach, using broad neural and imaging datasets to learn representations that transfer across subjects, tasks, and clinical conditions. However, many of these approaches are still trained on largely static objectives: reconstruction, masked prediction, or contrastive alignment. This approach often does not explicitly model the continuous-time process that generates observed neural activity. BrainDyn takes a complementary approach: rather than predicting masked tokens or reconstructing static observations, it learns by forecasting the continuous trajectory of neural activity over a structured brain graph, grounding representation learning in the dynamics themselves. BrainDyn is a structure-aware, dynamics grounded pretraining framework for an \textit{in silico} generative foundation model of the brain, analogous in spirit to virtual-cell models in biology.

\section{Graph construction ablation}
\label{app:spatial_connectivity_results}

Here we evaluate the effect of different graph construction strategies on BrainDyn's forecasting performance. We consider three approaches. 1) Spatial connectivity (the method used in the main text) constructs the graph  by computing the pairwise Euclidean distances between the physical locations of the nodes and connecting each node to its $k$ nearest neighbors.  2) Functional connectivity constructs a correlation graph by computing the Pearson correlation between the time-series signals of each pair of nodes. 3) Granger-causality~\citep{geweke1982measurement} is a common statistical measure of whether one time series is useful for predicting on another. We create a directed edge $(i,j)\in E$ with weight $w_{ij}=G_{i\rightarrow j}$, the Granger causality score which measures the effect of $x_i(t)$ on $x_j(t)$ (computed by taking the logarithm of the $F$ statistic obtained by running a regression to predict $x_j(t)$ based on $x_i(t)$).  (We note that, despite the name, Granger causality does not establish a causal effect, but merely measures whether knowledge of one quantity increases the ability of a regression model to predict another.)

\begin{table}[h]
\centering
\caption{Prediction of short-term and long-term horizons on resting-state fMRI dynamics across different graph types. Metrics reported as mean $\pm$ std across 5 randomized train/test/validation splits. Lower is better. Best results are \textbf{bolded}; second best are \underline{underlined}.}
\label{tab:graphcomparison_fmri}
\resizebox{\textwidth}{!}{%
\begin{tabular}{lcccccc}
\toprule
& \multicolumn{3}{c}{Short} & \multicolumn{3}{c}{Long} \\
\cmidrule(lr){2-4} \cmidrule(lr){5-7}
Graph Type & MSE & MAE & DTW & MSE & MAE & DTW \\
\midrule
Functional Connectivity 
& $\underline{0.0695 \pm 0.0096}$ 
& $\underline{0.1438 \pm 0.0117}$ 
& $\underline{0.0686 \pm 0.0050}$ 
& $0.6958 \pm 0.0815$ 
& $0.5688 \pm 0.0339$ 
& $0.2813 \pm 0.0175$ \\

Granger 
& $0.0854 \pm 0.0216$ 
& $0.1674 \pm 0.0211$ 
& $0.0793 \pm 0.0090$ 
& $\underline{0.6592 \pm 0.0310}$ 
& $\underline{0.5508 \pm 0.0143}$ 
& $\underline{0.2721 \pm 0.0073}$ \\

Spatial 
& $\mathbf{0.0531 \pm 0.0091}$ 
& $\mathbf{0.1238 \pm 0.0121}$ 
& $\mathbf{0.0599 \pm 0.0054}$ 
& $\mathbf{0.6374 \pm 0.0465}$ 
& $\mathbf{0.5356 \pm 0.0157}$ 
& $\mathbf{0.2654 \pm 0.0084}$ \\
\bottomrule
\end{tabular}%
}
\end{table}

\begin{table}[h]
\centering
\caption{Prediction of short-term and long-term horizons on resting-state EEG dynamics across different graph types. Metrics reported as mean $\pm$ std across 5 randomized train/test/validation splits. Lower is better. Best results are \textbf{bolded}; second best are \underline{underlined}.}
\label{tab:graphcomparison_eeg}
\resizebox{\textwidth}{!}{%
\begin{tabular}{lcccccc}
\toprule
& \multicolumn{3}{c}{Short} & \multicolumn{3}{c}{Long} \\
\cmidrule(lr){2-4} \cmidrule(lr){5-7}
Graph Type & MSE & MAE & DTW & MSE & MAE & DTW \\
\midrule
Functional Connectivity 
& $\underline{0.4131 \pm 0.0217}$ 
& $\underline{0.4264 \pm 0.0127}$ 
& $\underline{0.1935 \pm 0.0057}$ 
& $\underline{0.7607 \pm 0.0812}$ 
& $\underline{0.6128 \pm 0.0286}$ 
& $\underline{0.3068 \pm 0.0144}$ \\

Granger 
& $0.4411 \pm 0.0431$ 
& $0.4433 \pm 0.0286$ 
& $0.2023 \pm 0.0142$ 
& $0.7664 \pm 0.0674$ 
& $0.6188 \pm 0.0297$ 
& $0.3110 \pm 0.0149$ \\

Spatial 
& $\mathbf{0.4001 \pm 0.0253}$ 
& $\mathbf{0.4209 \pm 0.0126}$ 
& $\mathbf{0.1905 \pm 0.0061}$ 
& $\mathbf{0.7199 \pm 0.0292}$ 
& $\mathbf{0.6005 \pm 0.0161}$ 
& $\mathbf{0.3008 \pm 0.0080}$ \\
\bottomrule
\end{tabular}%
}
\end{table}

Tables~\ref{tab:graphcomparison_fmri} and \ref{tab:graphcomparison_eeg} show that across both datasets and both modes of forecasting, the spatial connectivity graph consistently achieves the lowest MSE, MAE, and DTW, while functional connectivity generally provides the second-best performance. Granger-causality based connectivity performs less favorably than the other graph constructions across most settings.

\section{Learned Positional Embeddings}
\label{app:pe}

\begin{figure}[h]
    \centering
    \includegraphics[trim=0mm 0mm 0mm 22.1mm, clip, width=1.00\textwidth]{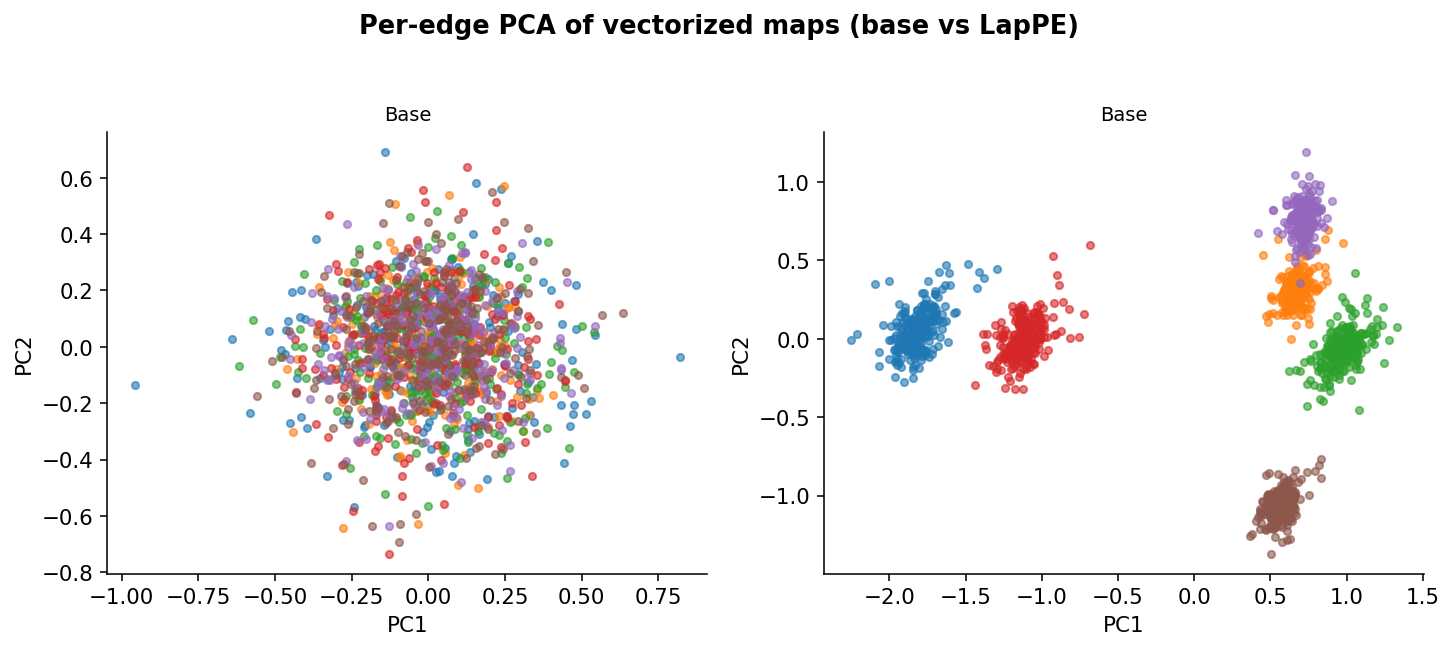}
    \caption{ Each color identifies one edge. Given a positional embedding the restriction maps exhibit clusters and distinct identities (right panel), unlike the restriction maps without positional embeddings (left panel).}
    \label{fig:pe}
\end{figure}

Here, we wish to justify the use of learned positional embeddings as a concatenated input to the MLP. These learned positional embeddings are used to give distinct node identities and allow the MLP to be able to deliver more individualized and better restriction maps, since the base LSTM embeddings are close together and not node-specific. Since the LSTM embeddings are so close to one another, it is hard for the model to give unique restriction maps for a given region. Here, we will seek to illustrate this visually. We vectorize the restriction maps and perform PCA. We color by edge identity and take a subsample for visual neatness (see Figure \ref{fig:pe}).

We choose to use learned positional embeddings rather than using other positional embedding strategies, such as Laplacian Positional Encoding (LapPE), because we do not require permutation equivalence or other more flexible quantities: we train on atlases and defined graphs across our data modalities, and not using LapPE enables greater diversity of embeddings, especially where it most matters; between connected edges, which will almost certainly have similar LapPE embeddings but could have very different ways of routing information which justify different maps.

\section{Computational Complexity}
\label{app:cc}



First, we construct the edge set using spatial proximity by computing pairwise distances between the $N$ channels or regions of interest and connecting each node to its $k$ nearest neighbors, resulting in a $k$NN graph with $|E| = \mathcal{O}(Nk)$ edges. Computing the pairwise distances and selecting the nearest neighbors costs $\mathcal{O}(N^2)$.
We first use an $L_1$-layer
LSTM for each node's context window of length $T_c$, giving a cost of  $\mathcal{O}(NT_c(d+L_1d^2))$, where 
$d$ is the hidden dimension and $N$ is the number of nodes. The positional embedding encoding incurs a negligible cost of $\mathcal{O}(Nd_{pe})$ where $d_{pe}$ is the dimension of the positional encoding. The cost of calculating the MLP-parametrized restriction maps is $\mathcal{O}(|E|(2dH+L_2H^2))$, where $L_2$ is the number of layers, $H$ is the largest hidden dimension in the network, and $|E|$ is the number of edges. The sheaf Laplacian then operates on the encoded node representations: for each edge $(u,v)$ it applies learned restriction maps $\rho \in \mathbb{R}^{d\times d}$ to project into the shared edge space, computes the inter-node disagreement, and pulls back to node space, costing $\mathcal{O}(|E|d^2)$. If there are $K$ rounds of message passing, then the cost is $\mathcal{O}(K|E|d^2)$. Finally, the neural ODE relies on an $ L_3$-layer MLP of width $V$, evaluated at each of $4S$ calls required by the RK4 integrator over a horizon of $S$ steps, contributing $\mathcal{O}(SN(dV+L_3V^2))$. The time embedding vector of dimension $2S$ requires $2S$ evaluations for each of the $4S$ RK4 calls made costing $\mathcal{O}(S^2)$. The overall per-sample complexity is therefore 
$$\mathcal{O}\Big(NT_c(d + L_1d^2) + |E|(dH + L_2H^2 + Kd^2) + S^2 + SN(dV + L_3V^2)\Big)$$
This scales quadratically in the forecast horizon $S$ and linearly in graph size $|E| \propto N$ for sparse graphs or quadratically for dense graphs $|E| \propto N^2$ with sheaf message passing dominating.

\section{Datasets and Preprocessing}
\label{app:datasets}

\paragraph{Resting-state fMRI (PNC)}
Resting-state BOLD was acquired under the Philadelphia Neurodevelopmental Cohort (PNC) protocol (single-site imaging; resting-state EPI with TR $=3$~s in the cohort neuroimaging description). For this work, publicly released derivatives from the Reproducible Brain Charts (RBC) initiative were used: functional scans were processed with the Connectome Pipeline for the Analysis of Connectomes (C-PAC; \texttt{cpac\_RBCv0}), including standard volume realignment and nuisance regression as distributed in that release. Activity was summarized using a fixed atlas-based parcellation, producing one mean BOLD time series per region per run. The atlas we used was the Schaefer~2018 cortical atlas (400 regions, 17-network solution)~\citep{schaefer2018local} in MNI152NLin6ASym space. Nuisance effects were addressed with a 36-parameter regression strategy, as documented for the RBC CPAC outputs used here. Each scan is thus represented as a multivariate time series of dimension 400 over $T$ retained volumes after processing. The fMRI dataloader constructs windows over parcellated ROI time series with shape (400, 40): 400 ROI signals and 40 consecutive timepoints (32 context, 8 horizon). Using a stride between windows of 40, this yields 1,664 training windows and 356 windows each for validation and test (2,376 total windows). We compute $z$-score statistics per node based on the entire training set, and apply these statistics across all three splits to normalize the dataset.

\paragraph{Resting-state scalp EEG (LEMON)}
Resting-state EEG data were acquired from the MPI Leipzig Mind-Brain-Body database, specifically the Leipzig Study for Mind-Body-Emotion interactions or LEMON \citep{lemon}. The EEG data is of moderate density, with 62 channels per scalp. However, due to technical dropout, all subjects had effective channels $58\leq n \leq 61$, with one channel being uniformly dropped across subjects. Since the number of channels missing for any given subject was small, interpolation to 61 channels was performed using MNE \citep{mne}. We train on 202 subjects, taking 150 windows per subject, with window size determined by mode (32 context, 8 horizon for short, 32 context 32 horizon for long) for 30,300 subject-labeled total windows, which are then split by subject into train, val, and test. We compute $z$-score statistics per node based on the entire train set, and apply these statistics across all three splits to normalize the dataset.

\paragraph{NEST simulations.}
We generated synthetic neural activity using NEST~\citep{gewaltig2007nest} from directed small-world networks of $100$ \texttt{iaf\_psc\_alpha} neurons. Each simulation lasted $2{,}000$ ms with $0.1$ ms resolution. Recurrent edges followed a directed small-world graph with \texttt{small\_world\_k}$=8$ and rewiring probability \texttt{small\_world\_beta}$=0.1$, yielding $400$ directed edges. All neurons received a shared Poisson background drive at $1{,}000$ Hz, with uniform recurrent synaptic weight and delay. Spike trains were binned into $10$ ms windows, converted to firing rates, and smoothed with a Gaussian kernel of standard deviation $20$ ms. For perturbation experiments, we selected one neuron uniformly at random and silenced it over a contiguous intervention window sampled within the central portion of the trajectory, allowing the perturbation to propagate through the recurrent simulated graph. We saved the unperturbed and perturbed firing-rate trajectories, ground-truth adjacency matrices, graph seeds, bin edges, and perturbation metadata for $1{,}000$ independently sampled network instantiations.

\section{Experimental Setup and Hyperparameters}
\label{app:exp}

BrainDyn is trained using the AdamW~\citep{adamw} optimizer with an initial learning rate of $10^{-3}$ and weight decay $10^{-5}$. The learning rate is scheduled using ReduceLROnPlateau with a multiplicative decay of $0.5$, a patience of $3$ epochs, and a minimum learning rate of $10^{-6}$. The restriction map MLP is initialized using Xavier uniform weights, scaled by $\gamma=0.1$, so the restriction maps begin near the zero matrix. $\mathcal{L}_F$ is quadratic in the restriction maps, so it starts approximately zero, which makes the message passing, defined in Eqn. \ref{eqn: sheaf MP} as $I-L_F$, begin near the identity before the maps are learned. Initial dynamics start near a graph-free model, from which sheaf structure is learned. The ODE is integrated using a fixed-step fourth-order Runge-Kutta (RK4) solver with step size $\Delta t = 1$, implemented via torchdiffeq~\citep{torchdiffeq}. Models are trained for up to $120$ epochs with a batch size of $64$, and we retain the checkpoint with the lowest validation loss. 
All experiments are run with automatic mixed precision (fp16) on a single NVIDIA H200 GPU.
As noted in Section \ref{sec:experiments}, we utilize a 70/20/10 train/validation/test split and report the average test set performance over five runs.
Additionally, we also ensure that all data associated to a given subject is contained in the same portion of the data set. 


\section{Evaluation metrics}
\label{app:metrics}

Mean squared error and mean absolute error are the standard scalar summaries in time series forecasting. They measure the pointwise deviation between a predicted trajectory and the observed one at each time step, which makes them straightforward and easy to compare across methods. Relying on them alone, however, can be misleading when the goal is to evaluate whether a model captures the local dynamics of the signal rather than just its mean level. A forecast that is slightly shifted forward or backward in time can incur a large MSE or MAE even if it reproduces the shape, frequency content, and transient structure of the signal almost exactly. Conversely, a prediction that misses those features but stays close in absolute value at every step may score well.

Dynamic time warping is a metric that alleviates this issue. Rather than aligning series at fixed time indices, DTW finds an optimal warping path between them that allows one series to be stretched or compressed locally to match the other, penalizing the cost of alignment along the way. This makes the resulting distance sensitive to differences in shape and local timing while being tolerant of small temporal offsets, which is precisely the kind of comparison that pointwise metrics and correlation miss. We report a normalized DTW distance throughout, defined as the total path cost divided by path length, so the score is comparable across trajectories of different lengths. 

\section{Baseline methods}
\label{app:baselines}

\paragraph{CNN-LSTM}
The common hybrid design in multivariate forecasting is convolutional feature extraction over time followed by a recurrent readout~\citep{agga2022cnnlstm, alshembari2025autoregressive}. We followed that design and implemented a per-node CNN-LSTM: each node's context trace passes through a shallow temporal convolutional stack and an LSTM, and future steps are predicted using learned step embeddings. This baseline captures local temporal structure but ignores brain connectivity.

\paragraph{EvolveGCN}
EvolveGCN~\citep{egcn} maintains node states on the graph while allowing the GCN weights to evolve as a function of the temporal context, yielding a standard dynamic-graph baseline that uses connectivity without a continuous-time latent integrator.

\paragraph{BIOT}
We use the public BIOT encoder~\citep{yang2023biot}, a transformer over short-time Fourier tokens of multichannel waveforms, with a lightweight forecasting head on the pooled embedding. On EEG the encoder sees the multichannel context window directly; on parcellated fMRI we expand each TR into a constant pseudo-waveform block so the spectrogram front end receives a compatible time axis.

\paragraph{ODEBRAIN}
We compare to the ODEBRAIN framework~\citep{jia2026ODEBRAIN}, which fuses spectral graph features with raw temporal information to form an initial state and integrates a learned neural ODE field to forecast evolving graph-structured representations of EEG. 

\paragraph{RiTINI}
RiTINI~\citep{bhaskar2024inferring} combines space-time graph attention with graph neural ODEs for dynamical systems on a graph prior. Our baseline preserves that decomposition (temporal encoding, graph attention on the FC edges, then ODE integration in latent space) but is reimplemented for batched $(\mathrm{batch},\mathrm{time},\mathrm{node})$ tensors and our FC graph construction, isolating the role of sheaf-structured coupling relative to attention plus continuous-time propagation on the same topology.

\paragraph{BrainLM}
BrainLM~\citep{ortega2024brainlm} uses a parcellation-based tokenization mechanism transforming regional activity into discrete tokens. The tokens are processed through a transformer-based masked autoencoder with an attention mechanism for capturing long-range dependencies. To align with our experimental paradigms, we evaluate its core architecture only.  

\paragraph{VAR}
Vector Autoregression is a classical statistical technique that generalizes simple linear regression to multiple dimensions and time series data. The method serves as a simple linear baseline modeling the forecast of a node as a linear combination of all nodes' past states.

\section{Stretching and Rotation of Restriction Maps}
\label{app:stretch-rotation}

\noindent
\begin{minipage}[t]{0.53\textwidth}
The anisotropic stretch is another measurement of this, with anisotropy a measure of how unevenly distributed the singular values are.
From the SVD $\rho = U \Sigma V^{\top}$, we can calculate the polar decomposition $\rho = Q P$, $ Q = U V^{\top}$, $ P = V \Sigma V^{\top}$, with $Q$ the rotation portion and $P$ the symmetric positive-definite stretching portion of the matrix. We quantify the rotational content of each map as the share of its deviation from the identity compared to the stretching factor,
\begin{equation}
  \text{rotation fraction}
  = \frac{\lVert Q - I \rVert_{F}}
         {\lVert Q - I \rVert_{F} + \lVert P - I \rVert_{F}},
\end{equation}

\end{minipage}%
\hfill
\begin{minipage}[t]{0.43\textwidth}
    \vspace{-8pt}
    \centering
    \includegraphics[trim=458pt 15pt 10pt 46pt, clip, width=0.95\linewidth]{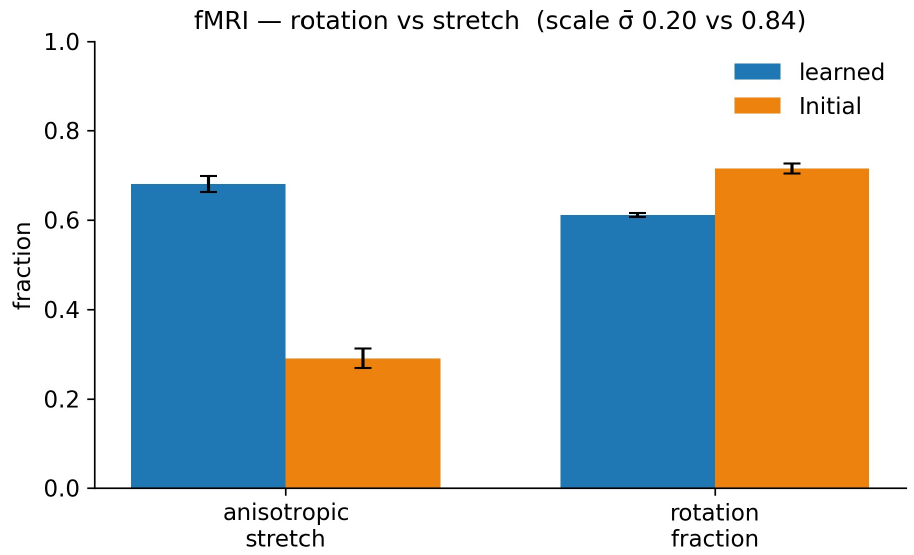}
    \captionof{figure}{Stretching fractions of the learned restriction maps compared to initialization.}
    \label{fig:stretchfraction}
\end{minipage}

so that the rotation fraction is 1 for a purely orthogonal map and 0 for a pure stretch.

We see that the maps move more towards stretching compared to initial, but that a significant portion of the restriction maps is still devoted to aligning vector spaces.

\end{document}